\documentclass[runningheads]{llncs}
\usepackage{graphicx}
\usepackage{amsmath}
\usepackage[switch]{lineno}
\usepackage{xcolor}
\usepackage{color}
\usepackage{subfigure}
\usepackage{amssymb}
\usepackage[ruled,vlined]{algorithm2e}

\begin{document}
\title{KnowRU: Knowledge Reusing via Knowledge Distillation in Multi-agent Reinforcement Learning}
%
%
\author{Zijian Gao, Kele Xu, Bo Ding, Huaimin Wang, Yiying Li, Hongda Jia}
%
%
\institute{College of Computer,National University of Defense Technology
\email{gaozijian19@nudt.edu.cn, kelele.xu@gmail.com, dingbo@aliyun.com, \{hmwang, liyiying10, jiahongda17\}@nudt.edu.cn}}
\titlerunning{KnowRU: Knowledge Reusing via KD in MARL}
\maketitle              
%
\begin{abstract}
Recently, deep Reinforcement Learning (RL) algorithms have achieved dramatically progress in the multi-agent area. However, training the increasingly complex tasks would be time-consuming and resources-exhausting. To alleviate this problem, efficient leveraging the historical experience is essential, which is under-explored in previous studies as most of the exiting methods may fail to achieve this goal in a continuously variational system due to their complicated design and environmental dynamics. In this paper, we propose a method, named ``KnowRU'' for knowledge reusing which can be easily deployed in the majority of the multi-agent reinforcement learning algorithms without complicated hand-coded design. We employ the knowledge distillation paradigm to transfer the knowledge among agents with the goal to accelerate the training phase for new tasks, while improving the asymptotic performance of agents. To empirically demonstrate the robustness and effectiveness of KnowRU, we perform extensive experiments on state-of-the-art multi-agent reinforcement learning (MARL) algorithms on collaborative and competitive scenarios. The results show that KnowRU can outperform the recently reported methods, which emphasizes the importance of the proposed knowledge reusing for MARL.

\end{abstract}
\section{Introduction}
Recently, reinforcement learning (RL) has made great progress to solve complicated tasks such as Atari games \cite{mnih2015human}, board games \cite{tesauro1995temporal}, video game-playing \cite{mnih2015human}, etc. With the compelling performance of single-agent models, multi-agent reinforcement learning (MARL) tasks, such as the collaboration and competition among multiple agents, have attracted increasing interests in several fields  \cite{vinyals2019grandmaster,zhou2018many} as the applications of MARL seems to be evident.

Current MARL algorithms are still highly task-specific and lack the ability to generalize to new environments. Moreover, for the resource-limited embedded systems, training the MARL system from scratch would be extremely time-consuming and resource-exhausting due to the huge complexity. Efficiently transferring and using knowledge between tasks can alleviate aforementioned issues, and sustainable efforts have been made in this fields. One category of the solutions employs the transfer learning paradigm to reuse the knowledge of historical experience, which can relief the burden of training a new model with previous experience \cite{da2019survey}. 

However, most existing transfer learning methods for multi-agents mainly depends on the hand-coded design, which requires the knowledge from the domain experts. 
For example, a method based on Pepper algorithm \cite{crandall2012just} proposed by \cite{hernandez2017towards} is used to obtain strategies for opponents in the adversarial scenarios and calculate policies against them. Then, agents in new scenarios judge the opponents' strategy and reuse the knowledge by learning from the calculated policy. Similarly, a genetically programming approach \cite{kelly2015knowledge} utilizes strategies from trained networks to new tasks. A set of neural networks are trained to predict the value of each action and obtain a set of action strategies to build a multi-tiered architecture for agents to learn when to trigger which strategy in new tasks. But, these approaches mainly relies on the hand-coded design which is closely related to the specific tasks, such as the mapping of state variables or modeling the opponents. Obviously, due to the difficulties in professional and hand-coded design for tasks, existing methods which require sufficient domain knowledge are not universal enough and difficult to deploy. It is desirable for knowledge reuse from previous experience using the method which can be widely used and easily deployed without considering how and what to transfer.

In this paper, we propose a method named KnowRU for knowledge reusing, which can be easily deployed in MARL algorithms. KnowRU can accelerate the training phase for new tasks, while improving the asymptotic performance of agents. Our motivation comes from knowledge distillation \cite{hinton2015distilling} which has been successfully applied in the computer vision field. Leveraging the knowledge of well trained agents in previous tasks as the historical experience, we employ the knowledge distillation approach to transfer the knowledge to agents for new tasks. Here, we suppose the action taken by the agent according to the environment is the simplest form of knowledge and action is determined by the output of the network. So, mimicking the output is a potential feasible way to reuse knowledge of historical agents. During the training phase of new tasks, agents are not only to get higher rewards in the environment but also to mimic historical agents' outputs. In this way, agents are able to learn from varying rewards and derive knowledge via mimicking the historical agents. The knowledge from historical agents can be viewed as a fundamental consensus among different tasks due to the discrepancy of tasks. Empirical experiments are conducted on different tasks and different MARL algorithms to validate the effectiveness of KnowRU. An example is illustrated in Figure 1 where agents must cooperate to arrive at the closest target as soon as possible. We retain the agents which are well trained in the past task, and then agents in the target task can observe how the well trained agents would work in the same situation and transfer knowledge from mimicking the observed actions.
\begin{figure}[h]
  \centering
  \includegraphics[width=\textwidth]{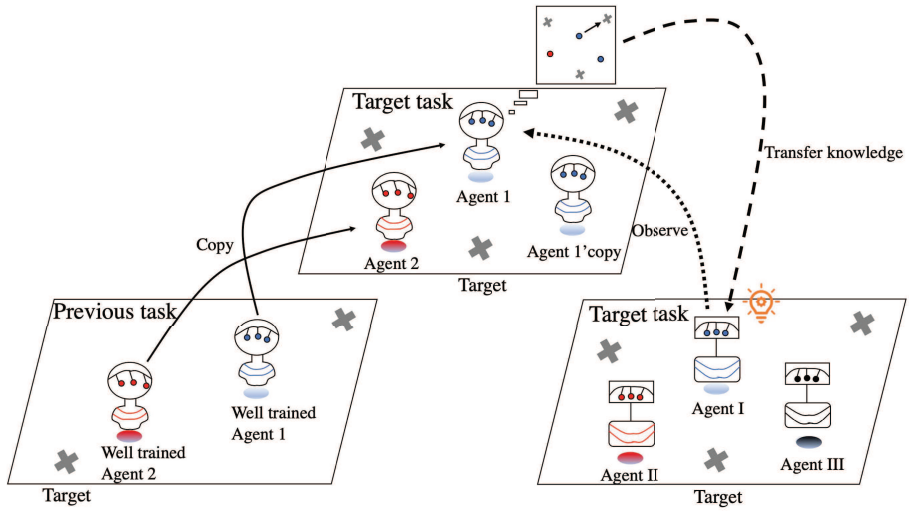}
  \caption{Overview of KnowRU. Agents in the target task mimic historical agents' actions in the same situation, and then effectively reuse the knowledge in the previous task and allow further learning in new environment.}
  \label{fig:example}
\end{figure}

The contributions in this paper are as follows:
\begin{itemize}
	\item We propose a task-independent knowledge distillation framework for MARL, which focuses on not only accelerating the training phase but also improving the asymptotic performance for new tasks. 
	\item Different strategies are explored to further improve the knowledge transfer performance.
	\item Extensive empirical experiments demonstrate the effectiveness of KnowRU in different task scenarios with different MARL algorithms.
\end{itemize}

\section{Related Work}
Multi-agent System (MAS) can be defined as the Stochastic Games (Markov Games, SG)  \cite{bowling2000analysis} which extends from basic Markov Decision Processes (MDP) . Based on the theories, many excellent algorithms of MARL have been proposed in past years. The traditional approaches developed for basic MDP such as Q-Learning \cite{watkins1992q} and policy gradient \cite{silver2014deterministic} fail to train agents well in MAS, because they have no head for considering environmental dynamics due to non-stationary policies of other agents. Instead, the MARL algorithms, such as MADDPG \cite{lowe2017multi} based on Actor-Critic architecture \cite{konda2000actor} takes all agents' information into account with a centralized critic. Furthermore, MAAC \cite{iqbal2019actor} learns a centralized critic with an attention mechanism to help agents focus on the vital information.

However, due to the huge sample complexity of traditional RL methods, it is tough to train agents from scratch every time specially in a continuous variational task, which would cost a lot of time and computational resource. The transfer learning \cite{taylor2009transfer} method is a practical way to alleviate the problem via the knowledge reusing. The vanilla MARL algorithms only focus on the learning process of the current task. There have been some methods aiming at mapping relationships among tasks. For example, \cite{koga2014stochastic} uses the relations among tasks to aggregate specific strategies in source tasks and generate one abstract policy which is used to help agents quickly adapt to new tasks. But the indescribable relations among tasks are the difficulty of abstracting policy. Using evolutionary algorithms to transfer knowledge in evolved multi-agent system is also a feasible way. \cite{didi2016multi} presents a neuro-evolution method that codifies the agents' policies through a neural network and optimizes the network's topology and weights though interactions with the environment. However, it relies heavily on humans to define the parameters and mapping between tasks. These transfer learning approaches mentioned above differ from ours in the following ways: (1) They mainly focus on how and what to transfer between tasks. (2) The nifty artificial design for specific tasks based on sufficient domain knowledge is the key to their success. In this paper, KnowRU only requires the policy model which is related to target tasks without considering about model structure, relationships between tasks and task-specific design. Compared to them, KnowRU shows wider application prospects and can be aggregated into more MARL algorithms in a more convenient way.

Knowledge distillation (KD) is a kind of knowledge transfer methods, which is firstly proposed in \cite{bucilua2006model} and becomes famous after \cite{hinton2015distilling}. Knowledge distillation compresses the knowledge of large-scale complex models (\textit{teachers}) into small and efficient models (\textit{students}) to facilitate the deployment of models on insufficient computing resources' devices. The idea of Knowledge distillation which inspires us is to train the small student model with not only true labels but also soft targets provided by the well-performed large teacher model. Now, the main KD methods can be divided into three categories: Logits-based methods for learning the output layer, Feature-based methods for learning the hidden layers and Relation-based methods for learning the relations between network layers. It's necessary mentioning that there are also some works about the application of knowledge distillation in RL \cite{lai2020dual}\cite{wadhwania2019policy}. They mainly focus on making use of the agent-level knowledge to tackle the problems in a single RL task with knowledge distillation paradigm. Instead, we mainly work on the knowledge reusing in multi-agent tasks and solve the problem of agents' rapid adaptation to tasks' dynamic changes in experiments.

\section{Methodology}
\subsection{Preliminaries and Notations}
We begin with the definition of the Markov Decision Processes (MDPs) \cite{da2019survey} in MARL, which can be denoted as a tuple $<S, U, T, R1...n,\gamma>$. Here, $S$ is the state space, $U$ is the joint action space, $T$ is the state transition function, $R_{i}$ is the reward function of agent $i$, $\gamma$ is the discount factor and $n$ is the number of agents. The observation of agent $i$ in the current state $S_{i}$ is $O_{i}$, then agent takes action $A_{i}$ with the policy  ${\pi}_{\theta_{i}}$ and produces the next state according to $T$. Agent $i$ can get the rewards $R_{i}$ from environment according to state $S_{i}$ and $A_{i}$. Agents simply aim to take proper actions which lead to maximal total return $R=\sum_{t=0}^{T} \gamma^{t} R_{i}^{t}$ where $t$ is the time horizon.

\begin{equation}\label{eq:policy}
\nabla_{\theta} J\left(\pi_{\theta}\right)=\nabla_{\theta} \log \left(\pi_{\theta}\left(a_{t} \mid s_{t}\right)\right) \sum_{t^{\prime}=t}^{\infty} \gamma^{t^{\prime}-t} r_{t^{\prime}}\left(s_{t^{\prime}}, a_{t^{\prime}}\right).\end{equation}

\textbf {Actor-Critic(AC)}: In order to overcome high variance of policy gradient,
Actor-Critic \cite{konda2000actor} methods use a function as the critic to evaluate actions, and replace the return term of policy gradient with the value function. In RL, given the state and action, according to Bellman equation,  the critic function can be written as Equation 2 and estimate the returns so that the actor can be updated at every step. In this way, models could better be updated with less noise. AC framework laid a solid foundation for the later multi-agent algorithms. Let $\pi$ in Equation 2 be the policy of agent.

\begin{equation}
\label{eq:ACQ}
	Q^{\pi}\left(s_{t}, a_{t}\right)=
	\mathbb{E}_{r_{t}, s_{t+1} \sim E}\left[{r\left(s_{t}, a_{t}\right)+}\right. \\
	 \left. {\gamma \mathbb{E}_{a_{t+1} \sim \pi}\left[Q^{\pi}\left(s_{t+1}, a_{t+1}\right)\right]}\right].
\end{equation}

\textbf {Multi-Agent Deep Deterministic Policy Gradient (MADDPG):} MADDPG \cite{lowe2017multi} is an extension of Deep Deterministic Policy Gradient (DDPG) \cite{lillicrap2015continuous} which uses ``target+online'' networks and the experience replay to deal with  the failure of actor-critic in continuous action space. The critic in vanilla DDPG focuses only on the local information of current agent, not the global view, resulting in a not-so-stable performance in multi-agent system. Instead, MADDPG  exploits global observations and actions for all agents with a centralized action-value function: 
\begin{equation}\label{eq:MADDPG}
Q_{i}^{\pi}\left(\mathbf{x}, a_{1}, \ldots, a_{N}\right), \end{equation}
where $\mathbf{x}=\left(o_{1}, \ldots, o_{N}\right)$ , $i$ denotes the current agent and let $\pi=\left\{\pi_{1}, \ldots, \pi_{N}\right\}$ be the set of all agent policies. If all actions taken by all agents are accessible, the learning processes conform to the Markov Property. That's why MADDPG works well in multi-agent systems.

\textbf {Multi-Actor-Attention-Critic (MAAC):} 
MAAC \cite{iqbal2019actor} then makes a significant contribution to Actor-Critic-based MARL algorithm with the attention mechanism. Every agent queries information of others' observations and actions and then estimates its value with the information. The Q-value function $Q_{i}^{\psi}$ takes current agent $i$'s observation, action and other agents' contribution into consideration as: \begin{equation}\label{eq:MAAC} Q_{i}^{\psi}(o, a)=f_{i}\left(g_{i}\left(o_{i}, a_{i}\right), x_{i}\right),\end{equation}
where $f_i$ is a multi-layer perceptron (MLP), while $g_i$ is a MLP embedding function. The contribution coming from key-value memory model is a weighted value of other agents: \begin{equation}\label{eq:MAAC}x_{i}=\sum_{j \neq i} \alpha_{j} v_{j}=\sum_{j \neq i} \alpha_{j} h\left(V g_{j}\left(o_{j}, a_{j}\right)\right), \end{equation} where $v_j$ denotes a function of agent $j$’s embedding which is encoded with an embedding function and then linearly transformed by a shared matrix $V$, while $h$ is an element-wise nonlinearity and $\alpha_{j}$ represents the attention weight .

In this paper, we are in line with the Actor-Critic methods, and take MADDPG and MAAC as our baselines.
\subsection{Knowledge Distillation}
As mentioned above, because the teacher model provides more useful information for the student model, KD has achieved success in the field of computer vision. The soft probabilities output of trained teachers is the key of distillation. Let $a_{t}$ be the input logits of final softmax layer of teacher network where $a_{t}= [a_{1},a_{2},....,a_{j}]$. The logits are converted into probabilities $q_{t}= [q_{1},q_{2},....,q_{j}]$ with softmax function: $q_{i}=\frac{e^{a_{i}}}{{\Sigma}_{j} e^{a_{j}}}$. In order to extract more information compared with true labels, \cite{hinton2015distilling} proposes to soften the teacher probabilities with temperature $T$:
\begin{equation}\label{eq:soft}
q_{i}=\frac{exp(a_{i}/T)}{{\Sigma}_{j} exp({a_{j}/T})}.
\end{equation}

In KD, such \textit{dark knowledge} from soft output of teacher provides more information than true labels. Based on the same image input $x$, the teacher network and student network produce probability $q_{t}(x)$ and $q_{s}(x)$ with Equation~(\ref{eq:soft}). The gap between $q_{t}(x)$ and $q_{s}(x)$ is usually penalized by Kullback-Leibler divergence (Equation~(\ref{eq:K})) or cross-entropy loss :
\begin{eqnarray}\label{eq:KL}
\mathcal{L}_{K D}=T^{2} K L\left(q_{s}(x),q_{t}(x)\right).
\end{eqnarray}
\begin{equation}\label{eq:K}
KL(P \| Q)=\sum P(x) \log \frac{P(x)}{Q(x)}
\end{equation}
where P(x) and Q(x) are two probability distributions on random variable $x$. Temperature $T$ in Equation~(\ref{eq:KL}) also aims to soften the output of the teacher network. Then the student network could reuse knowledge the teacher network by the back propagation of ${L}_{KD} $. Knowledge distillation inspires us that minimizing the gap between previous agents and current agents with the skill of distillation is the essence of knowledge reusing. We then draw upon such KD thought in our MARL research to reuse knowledge and verify the feasibility of KnowRU in section 3.3.

\subsection{Knowledge Reusing via Mimicking}
Imagine a real-world scenario where our training task is constantly variational and the number of agents is also increasing. It's impractical for us to train agents from scratch whenever task changes because of costly time and resource. We believe that there are connections between tasks, and the knowledge learned in previous tasks can be regarded as historical experience. So, knowledge reusing from historical experience is particularly important. However, we have argued that some existing works cannot reuse knowledge directly and effectively without exquisite design and expert-level experience. To solve the problems like rapid adaptation to dynamic changes on the number of agents, it's necessary to find an easier and more practical way of reusing knowledge. We aim to design a method that works well in such scenarios.

In this paper, we are inspired by KD and propose to reuse knowledge by mimicking to minimize the gap between tasks in MAS. We design a task-independent knowledge reusing method that can be applied based on multiple MARL algorithms without task-specific design. Firstly, we make use of the policy models of the well-trained agents which are homogeneous with the current training agents in previous task related to the current task. Then, we pair every current agent and every homogeneous policy model. If the number of current agents is greater than the number of policy models, repeated pairings are allowed. In the training process, same observations of current agent are input into both the previous policy model and the current model. Based on the same input, the logits which is the input of the last softmax layer is used to measure the gap between tasks with loss function. At the same time, current agents get the feedback from environment by returns. 

\begin{figure}[h]
  \centering
\includegraphics[width=\textwidth]{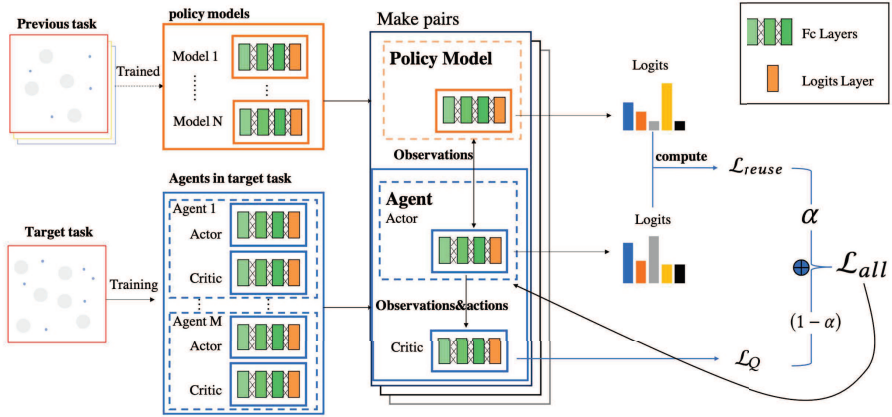}
  \caption{Workflow of KnowRU based on Actor-Critic algorithm. 1.Match the policy models in the previous task with the agents in the target task. 2.Calculate the loss from environment and the gap between Actor network and policy model. 3.While agent gets feedback from the environment, it also mimics the actions of the policy model.}
  \label{method}
\end{figure}

In this way, the agents of new task are trained by not only maximizing the total return from environment but also mimicking the well-trained previous policy models' output. At last, considering the difference between tasks, how to reasonably combine the mimicking gap loss and the returns from environment is also the key to the training. Here, we use hyperparameter $\alpha$ to adjust the relationship between these two factors and finally get the total loss $\mathcal{L}_{all}$ for back propagation. Figure~\ref{method} illustrates the main components of KnowRU based on Actor-Critic framework which is widely used in MARL. KnowRU has been proven feasible in a variety of experimental scenarios in section 4.

\textbf{Mimicking.}
In the computer vision field, the teacher model used to transfer knowledge is a well-trained model trained on the same dataset as the current student model. For this reason, in order to extract dark knowledge, it's necessary to soften teacher model's output probability with Equation~(\ref{eq:soft}) and minimize the gap between them with Cross-Entropy loss or KL loss using Equation~(\ref{eq:KL}). However, in MARL, the source task and target task might be different. The previous policy models may overconfident or not be authoritative in the new task. So, it's not necessary to soften the policy model's probability with hyperparameter $T$. Assume that $a_{p}$ and $a_{c}$ are the input logits of final softmax layer in previous policy model and current model, where $a_{p}= [a_{1},a_{2},....,a_{n}]$ and $a_{c}= [a'_{1},a'_{2},....,a'_{n}]$. Here, we use Mean-Square Error (MSE) loss as $\mathcal{L}_{reuse}$ loss function.
\begin{equation}\label{eq:MSE}
\mathcal{L}_{MSE}=\frac{1}{n} \sum_{i=1}^{n}\left({a}_{i}-a'_{i}\right)^{2}.
\end{equation}
\textbf{Task's guide and specialization.}
We have declared that the previous policy model and current model are usually working for different but similar tasks in MARL. The previous policy model cannot completely bootstrap the current learning, but there is still some knowledge between similar tasks that can be described as consensus. Some studies \cite{pan2009survey}\cite{weiss2016survey} show that unprincipled reuse of knowledge may not help training but hinder training. So, it's vital to reuse the consensus in a reasonable way. We naturally divide the training process into two phases: $phase\uppercase\expandafter{\romannumeral1}$  \textbf{guide} and $phase\uppercase\expandafter{\romannumeral2}$ \textbf{specialization}. In $phase\uppercase\expandafter{\romannumeral1}$, the previous policy model mainly guides the current model to reuse knowledge. As the training progresses, the agents enter into the stage of \textbf{specialization} stepwise, and the difference between tasks will gradually increase. We have shown in Figure 2 that the training depends on two factors, $\mathcal{L}_{reuse}$ and  $\mathcal{L}_{Q}$. We use the hyperparameter $\alpha$ to adjust the weight of the two factors and achieve the transition between the two stages, so
\begin{equation}\label{eq:Loss}
\mathcal{L}_{all}=\alpha \mathcal{L}_{reuse}+(1-\alpha) \mathcal{L}_{Q},
\end{equation} 
where $\alpha \in[0,1]$. Here $\mathcal{L}_{reuse}:=\mathcal{L}_{MSE}$,
and the $\mathcal{L}_{Q}$ is defined as the Q-value coming from action-value function in algorithms. Through the control of $\alpha$, we simulate the shift of focus from guide phase to specialization phase in the learning process. The $\alpha$ usually starts with a value greater arround 0.5 and drops linearly to 0.02 in our experiment settings. We believe that the low weight of knowledge reusing left behind can provide some noise for training to avoid overfitting to the task. It's worth mentioning that the $\mathcal{L}_{reuse}$ and  $\mathcal{L}_{Q}$ may not be of the same order of magnitude which may have adverse effects in RL. Experiments show that scaling $\mathcal{L}_{reuse}$ to an order of magnitude with $\mathcal{L}_{Q}$ is a wise solution for the problem.The algorithm of KnowRU based on AC framework is shown in Algorithm 1.
\begin{algorithm}[htb]
\caption{The Training Process based on AC framework}

\textbf{Initialization:} the parameters $\theta_{A}$,$\theta_{C}$ of student's actor, critic network, the parameter $\theta_{P}$ of previous policy model and the parameters  $\alpha$, $\beta$ of the weight of knowledge reusing and weight decay value.

\textbf{Output:} the parameter $\theta_{A}$ of  actor networks,the parameter $\theta_{C}$ of critic networks
for every agent:

\textbf{for} episode=1 to max-episodes:

\quad\textbf{for} step $i$=1 to max-steps-in-episode:

\quad\quad take action $a_i$ : $a_i=\pi_{\theta_{A}}(s_i)+\mathcal{N}_i$

\quad\quad get $r_i$ from environment, and observes new state $s_{i+1}$ 

\quad\quad store transition $(s_i, a_i, r_i, s_{i+1})$ into replay buffer

\quad \textbf{end for} 

\quad randomly sample N transitions from replay buffer

\quad \textbf{for} j=1 to N by step k: 

\quad\quad get $n_j$ samples by order 

\quad\quad get $logits_{P}$ by 
$\pi_{\theta_{P}}(s_i)$

\quad\quad get $logits_{S}$ by 
$\pi_{\theta_{A}}(s_i)$

\quad\quad compute $\mathcal{L}_{reuse} =\frac{1}{n_j} \sum_i (\mathcal{L}_{function}(logits_{P},logits_{S}))$
 
\quad\quad where $\mathcal{L}_{function}$ := $\mathcal{L}_{MSE}$

\quad\quad get $\mathcal{L}_{Q} = \frac{1}{n_j} \sum_i(Q_{\theta_{C}}(s_i, a_i))$

\quad\quad optimize actor network by minimizing: $\mathcal{L}_{all} =\alpha \mathcal{L}_{reuse}+(1-\alpha)\mathcal{L}_{Q}$

\quad\quad optimize critic network by minimizing: $\mathcal{L}_{critic} =\frac{1}{n_j} \sum_i|(Q_{\theta_{C}}(s_i, a_i)-r_i)|$

\quad\textbf{end for}

\quad\textbf{if} $\alpha > 0.02$:

\quad\quad $\alpha = \alpha - \beta$
 
\textbf{end for}
\end{algorithm}
\section{Experiments and Analysis}
\subsection{Experimental Setup}
\textbf{Multi-agent particle environment.} We construct three scenarios in multi-agent particle environment (MPE) \cite{lowe2017multi} to validate the performance of our method. The environment in MPE consist of $N$ agents and $L$ landmarks, and the relation among agents could be cooperation or competition. There are also some predefined scenarios including typical settings of cooperation (all agents maximize a shared total reward) and competition (different groups of agents have conflicting goals), we can also create more complex scenarios we need based on the engine of MPE. We test KnowRU in both cooperation and competition scenarios based on MADDPG and MAAC.

As mentioned above, knowledge reusing could play a vital role in continuous variational tasks. For this reason, we construct a series of environments to simulate the variations in such tasks by changing the number of agents and landmarks. In experiments, KnowRU can bring an inspiring impact on accelerating training and improving performance, mainly making comparisons among the results of basic algorithms and KnowRU. All contrasts are based on the same experimental conditions. The experiments consist of three scenarios involving cooperation and competition: simple\_spread, simple\_adversary and cooperative\_treasure\_collection.
\begin{figure}[h]
  \centering
  \includegraphics[width=\textwidth]{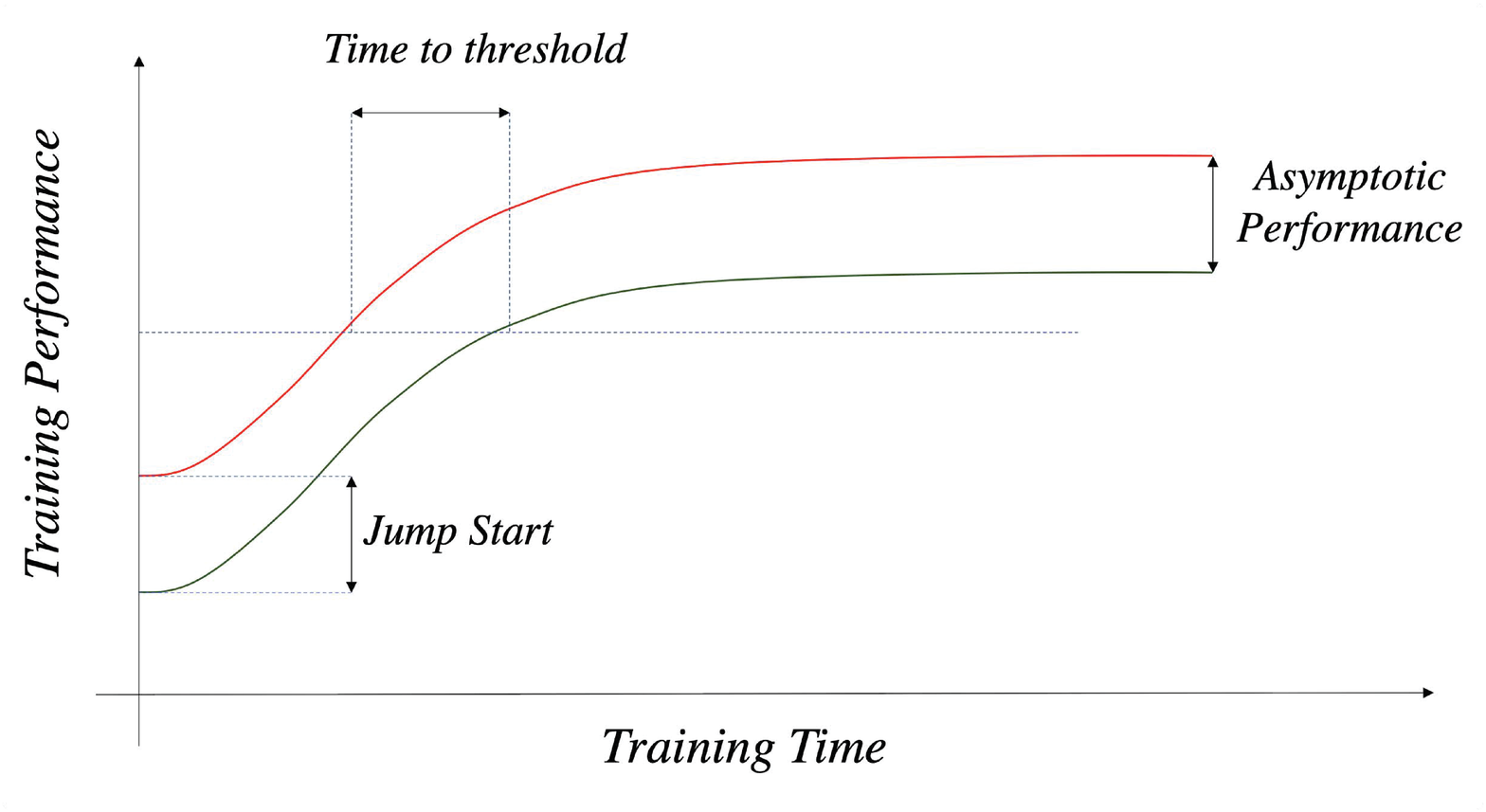}
  \caption{Transfer performance metrics.}
  \label{stand}
\end{figure}

\textbf{Performance Metrics.} There is also a common standard to measure the success of knowledge reusing in our experiments,  summarized by  \cite{taylor2009transfer} and resumed by \cite{da2019survey}, illustrated in Figure~\ref{stand}. The three main indicators are: (1). \textbf{Jump Start}: Measuring the improvement of performance at the beginning of training. (2). \textbf{Time to threshold}: For tasks which may get a significant result at some point, the learning time to reach the level is meaningful. (3). \textbf{Asymptotic Performance}: In complex tasks, agents might fail to reach the optimal performance and just reach a suboptimal one. Knowledge reusing might help agents to reach a higher performance and the before-and-after gap of performance is called asymptotic performance.

\subsection{Simple\_spread Scenario}
Simple\_spread is a predefined typical cooperative scenario of MPE. In this environment, agents must cooperate to reach a set of landmarks(targets) without communication. The targets are no difference, and agents need to collect the shared rewards by arriving at all targets as quick as possible. Meanwhile, agents are penalized for collision with each other. 

\begin{figure}
\centering
\subfigure[Tasks in simple\_spread scenario.]{\label{task1}
\includegraphics[width=0.47\textwidth]{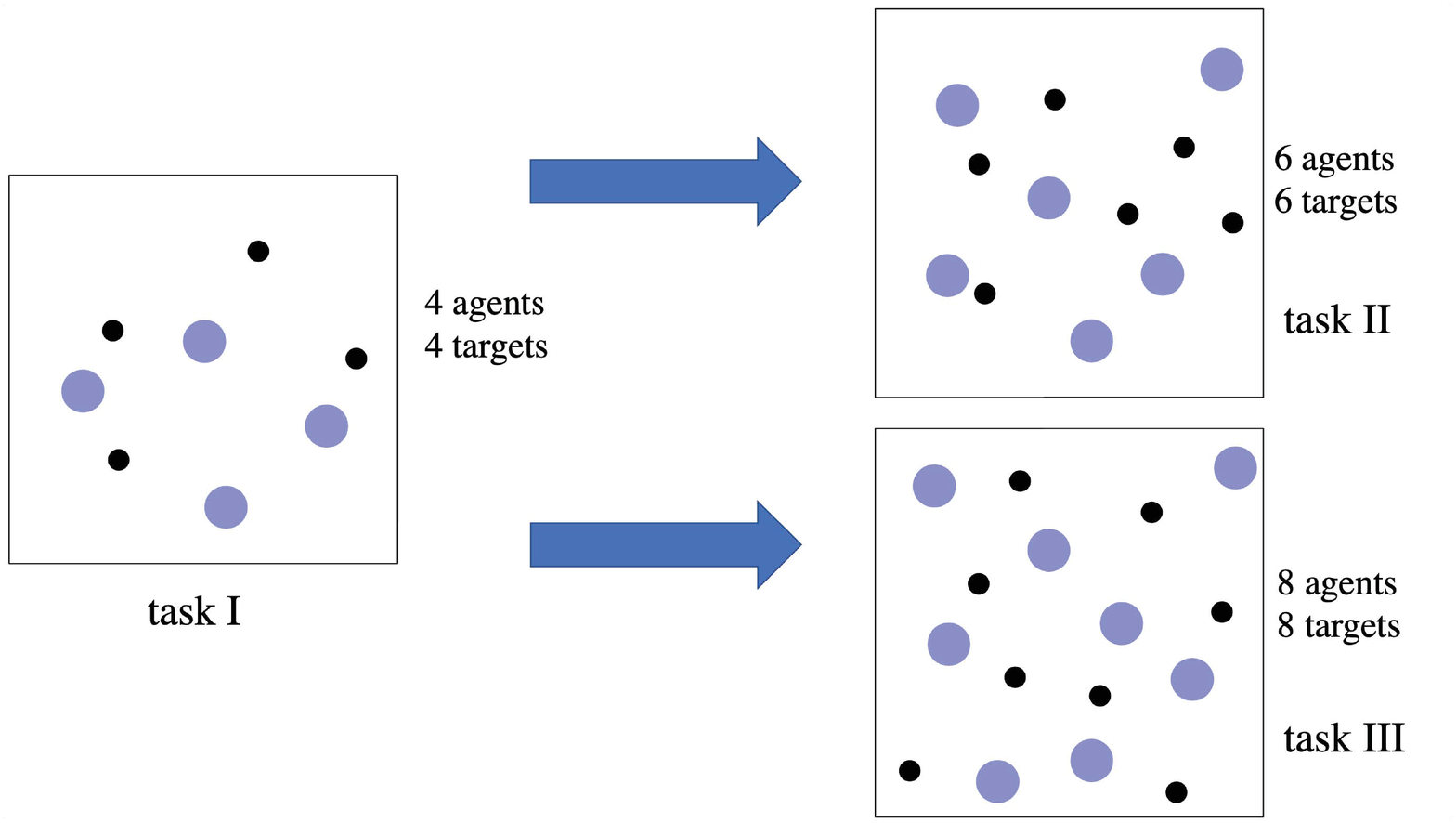}}
\hspace{0.01\linewidth}
\subfigure[Tasks in simple\_adversary scenario.]{\label{task2}
\includegraphics[width=0.47\textwidth]{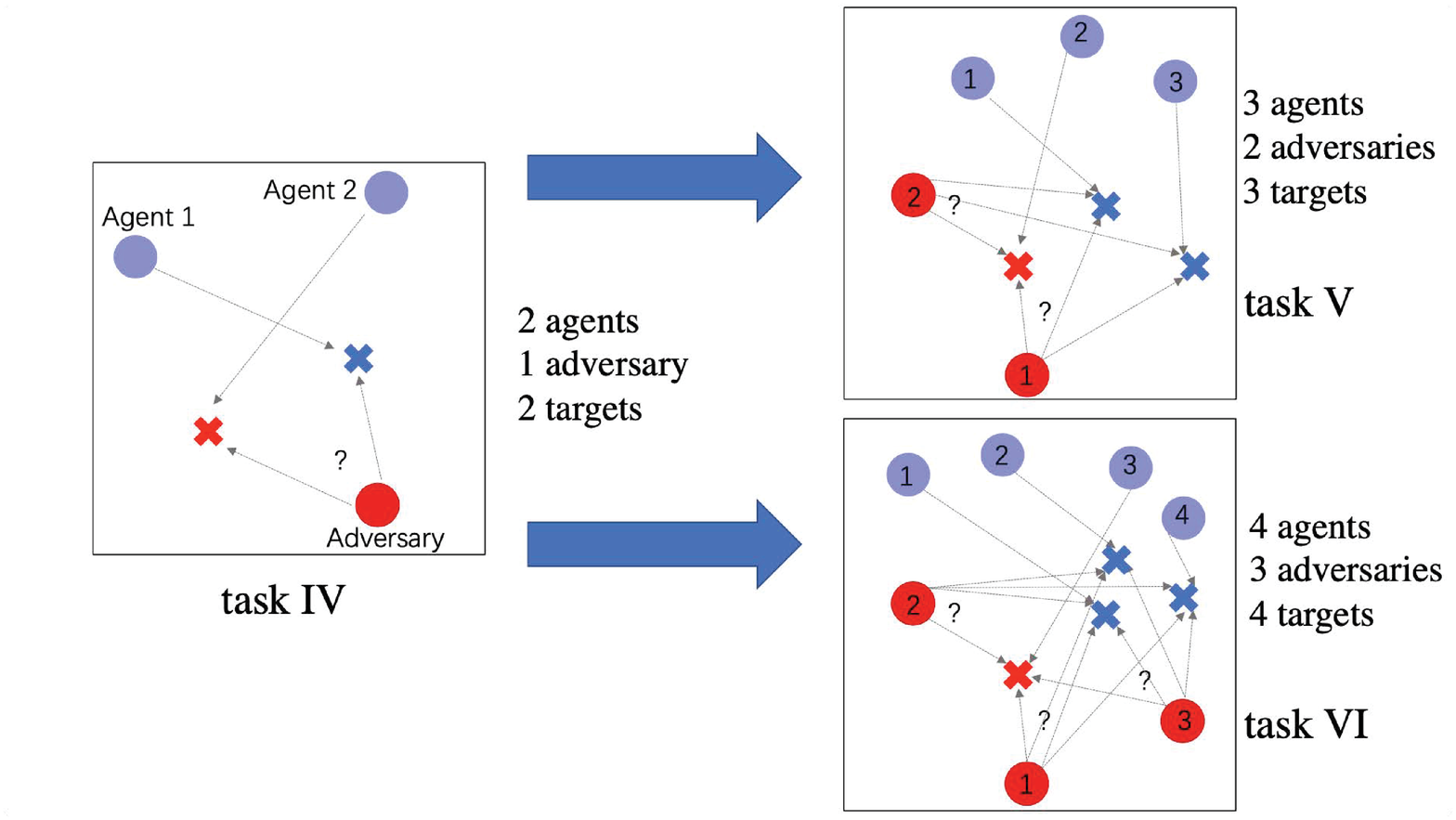}}
\caption{The tasks in Simple\_spread \& Simple\_adversary.}
\end{figure}

In our experimental settings, we take the task containing 4 agents and 4 targets as the previous policy models' training scenario, called task I. In a real-world scenario, with the continued variation of the training mission, the number of agents and targets usually changes. For this reason, we design two scenarios as current agents' training scenarios. One of them is constituted of 6 agents and 6 targets and the other is composed of 8 agents and 8 targets, called task II and task III respectively, as shown in Figure~\ref{task1}. In simple\_spread, we use MADDPG as the baseline algorithm and implement KnowRU based on it. In addition, we specially add a control group that initializes the current agent with the previous policy model, which is called "MADDPG with initialization", to verify the infeasibility of training directly based on the policy model. 

\subsection{Simple\_adversary Scenario}
Simple\_adversary is a predefined typical competitive scenario of MPE. In this scenario, $n$ agents must cooperate to reach one target landmark from total $n$ landmarks(targets) without communication. The agents need to maximize the shared rewards by minimizing the distance between the right target and one of agents which is closest to it. Meanwhile, the adversaries also want to reach the target without knowing which target is the right one and the agents are also penalized by the adversary distance to the target. Because agents know the correct target point, the agents have an advantage at the beginning. As the training progressed, adversaries learned how to distinguish the correct target, and achieve the balance of power.

In our experimental settings, we take the task which contains 2 agents, one adversary and 2 targets as the previous policy models’ training scenario, called task IV.  And we design two scenarios as current agents’ training scenarios. One of them is constituted of 3 agents, 2 adversary and 3 targets, the other is composed of 4 agents, 3 adversary and 4 targets, called task V and task VI, as shown in Figure~\ref{task2}. For this scenario, we only implement KnowRU with adversaries to help them distinguish the correct target at the beginning.
\begin{figure}
\centering
\subfigure[task II in Simple\_spread.]{\label{6}
\includegraphics[width=0.47\textwidth]{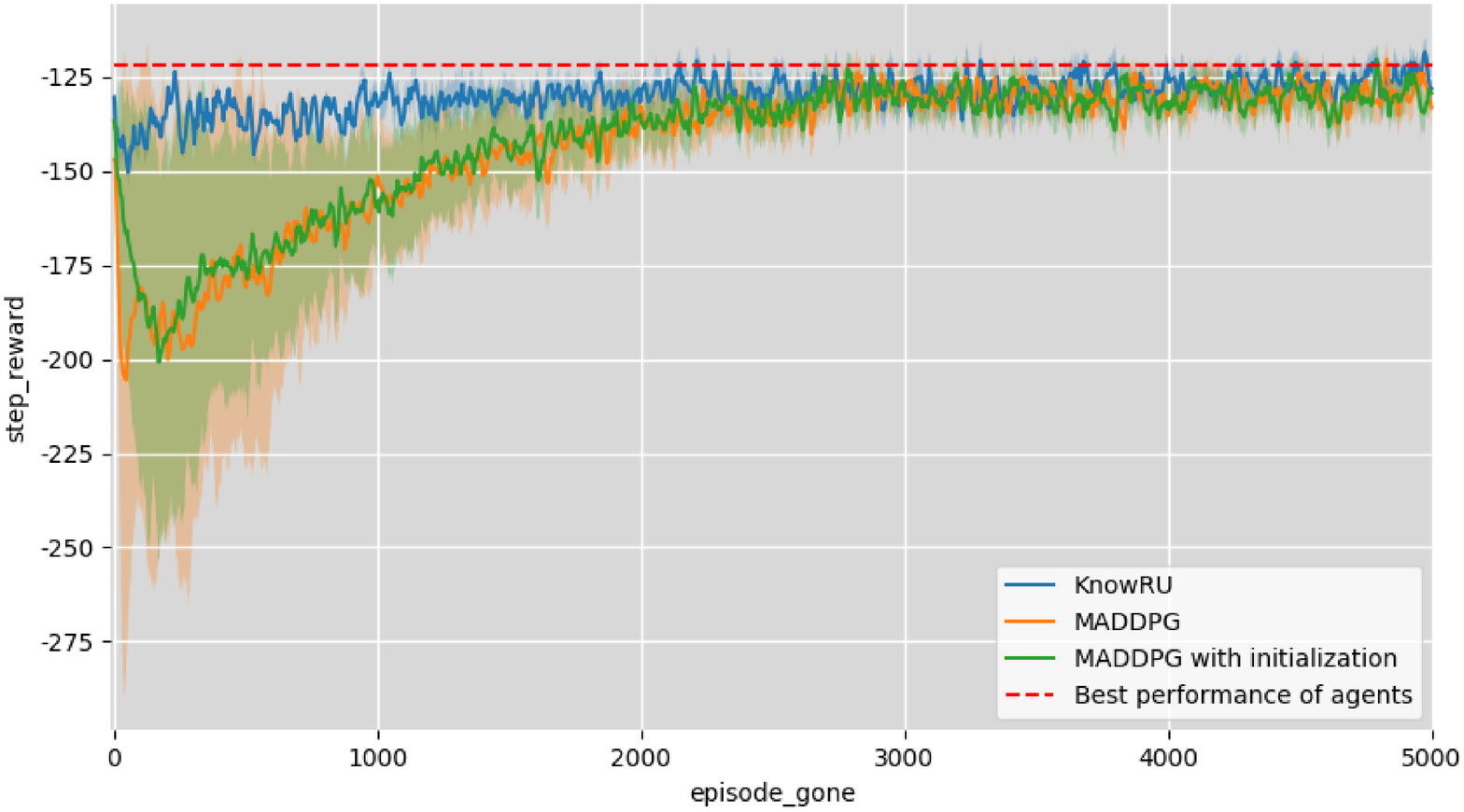}}
\hspace{0.01\linewidth}
\subfigure[task III in Simple\_spread.]{\label{8}
\includegraphics[width=0.48\textwidth,height=3.19cm]{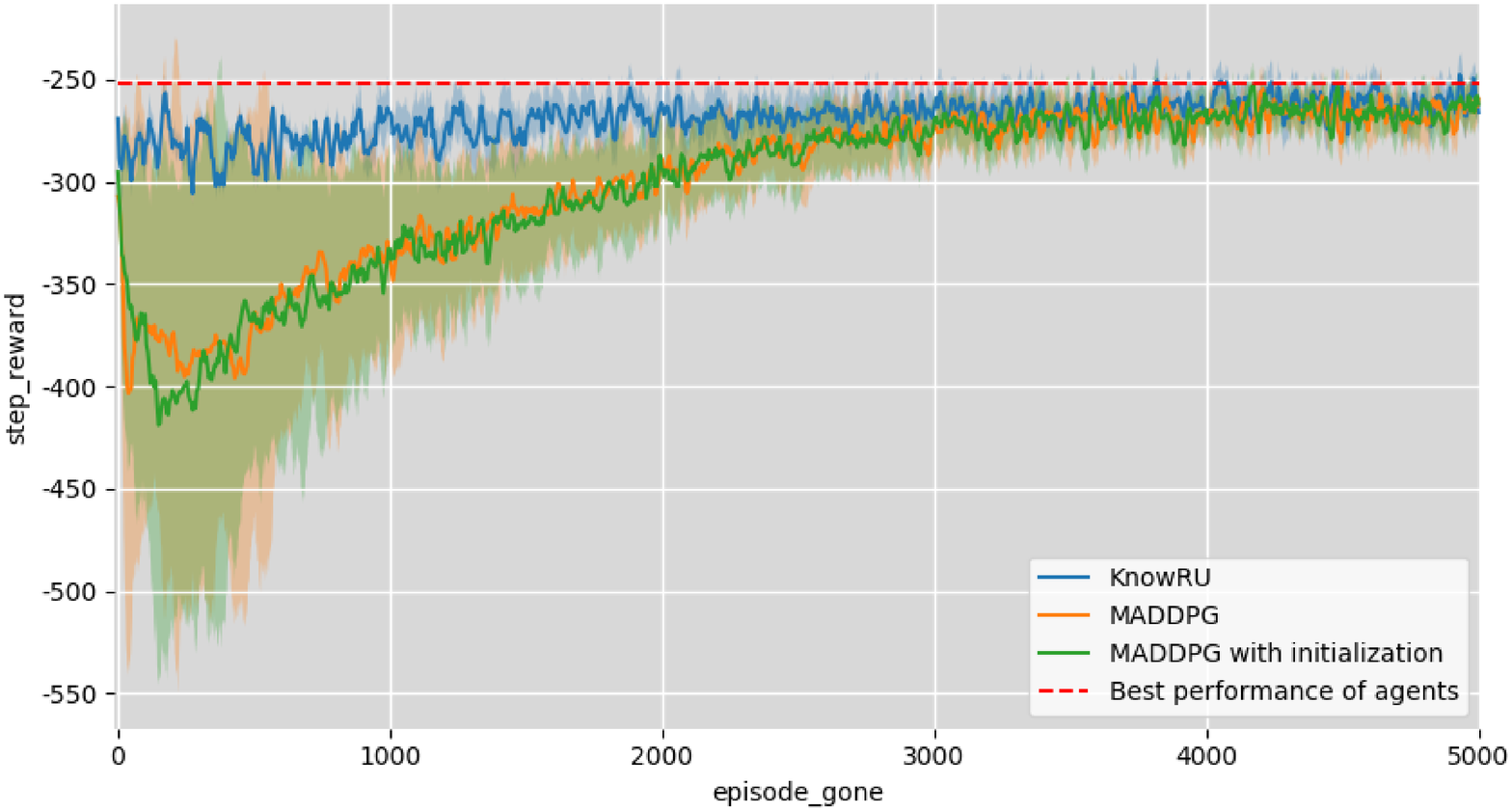}}
\vfill
\subfigure[task V in Simple\_adversary.]{\label{323}
\includegraphics[width=0.47\textwidth]{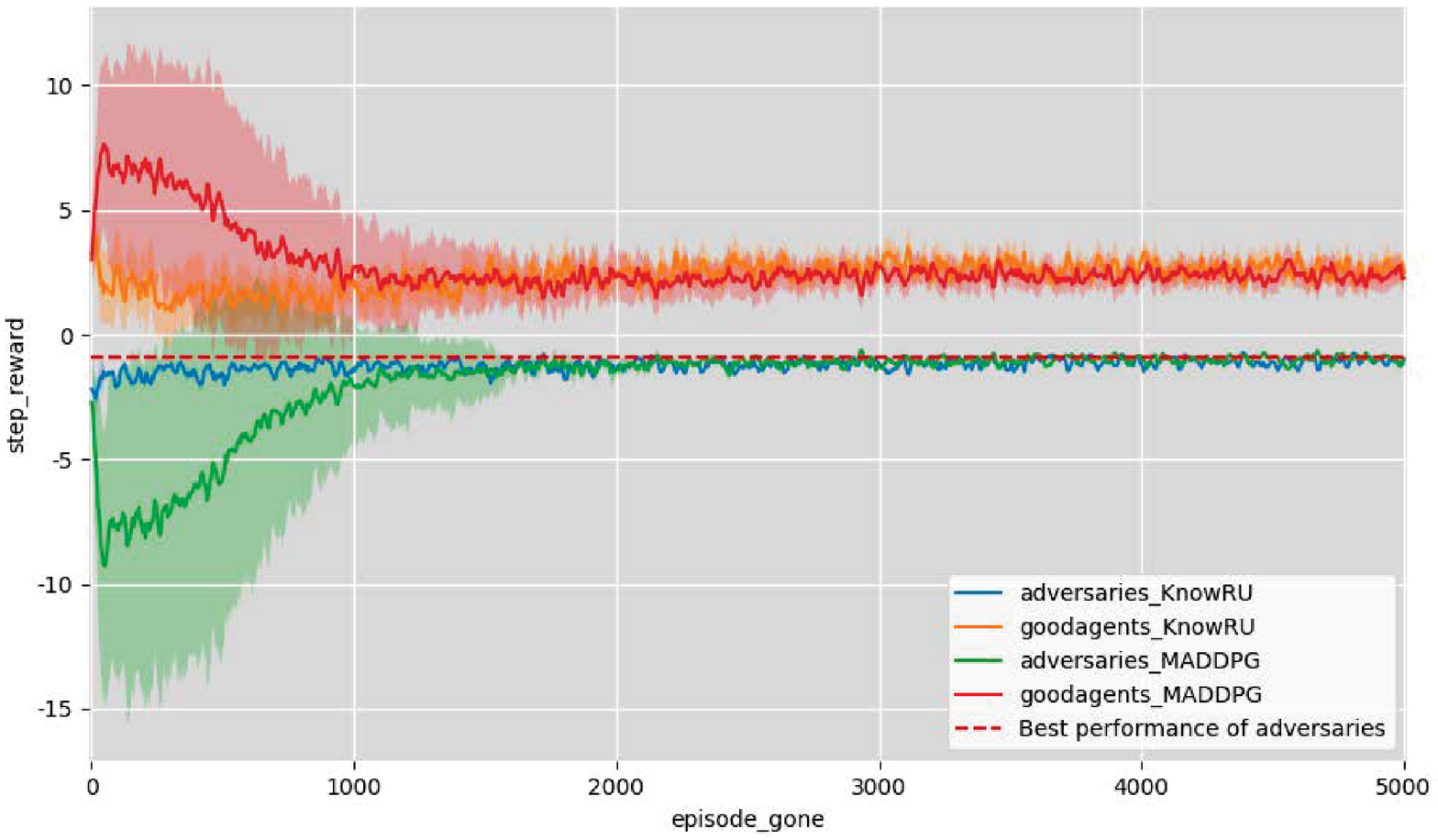}}
\hspace{0.01\linewidth}
\subfigure[task VI in Simple\_adversary.]{\label{434}
\includegraphics[width=0.48\textwidth,height=3.3cm]{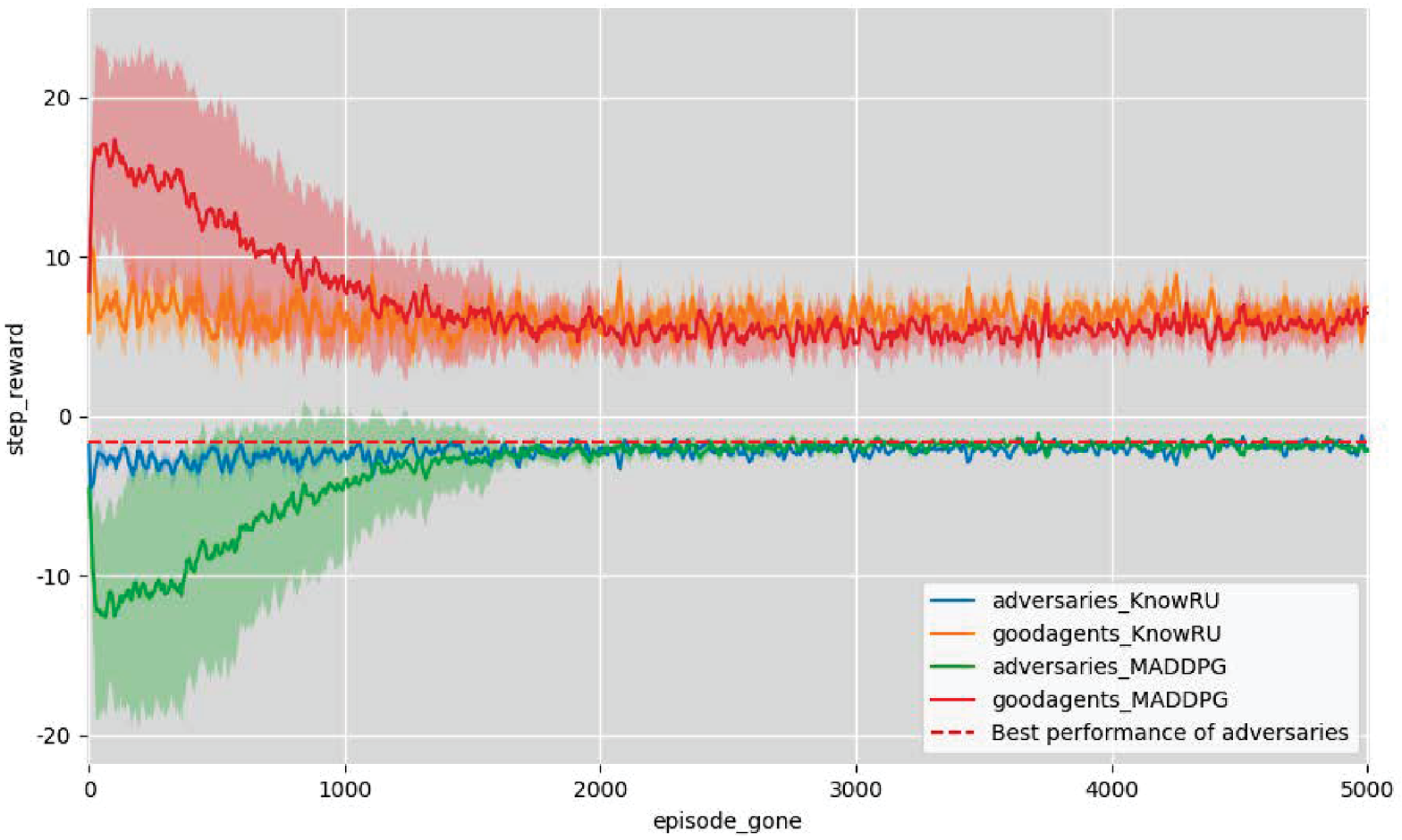}}
\caption{Knowledge reusing in Simple\_spread \& Simple\_adversary.}
\label{asubfig}
\end{figure}

In the first two scenarios, we take the average step reward from environments in each episode to measure the effect of training. The higher the reward, the better. The actor and critic networks in experiments are all randomly parameterized by a four-layer fully connected MLP with 64 units per layer. We set 5000 episodes to ensure the convergence, every episode consists of 25 steps and the networks update every 4 episodes. The hyperparameter $\alpha$ and $i$ are set to 0.5 and 0.02. All the results and 95\% confidence intervals (95\% CI) are illustrated in Figure~\ref{asubfig}. The red lines represent the best performance  in these scenarios.

\textbf{Discussion.} From the Figure~\ref{asubfig}, it's obvious that the performance of KnowRU in all experiments has been greatly improved at the beginning of training and the episodes required for training convergence are also greatly reduced, which satisfies two of the three indicators and prove that KnowRU successfully reuse knowledge. It's worth mentioning that when we train the initialized agents in task II \& task III, compared to MADDPG, the results did not clearly get better. This is caused by characteristic of models that they usual only work well in the scenarios they have been trained. When the task changes, the former models is no longer applicable due to over-fitting. Meanwhile, MADDPG shows great fluctuations in the training process without prior knowledge, especially in the initial stage of training, the performance decreased significantly. As for KnowRU, it's almost getting the convergence results at the beginning, achieving the goals of reusing knowledge in new tasks. Compared to MADDPG, KnowRU shows great performance and small fluctuations during training. We believe that the reason why KnowRU works is that it effectively narrows the solution space by providing more prior knowledge and thus narrows the space for exploration. We also find that during the training process, the fluctuation of the loss function value caused the unstable performance of the agents. When we use KnowRU to train agents, the oscillating amplitude of the loss value for backpropagation is apparently smaller and $\mathcal{L}_{reuse}$ first decreases and then increases, which proves the effectiveness of the two-phrase design of KnowRU. The reason why KnowRU does not appear to learn beyond its initial performance is that agents have reached the best performance in the scenarios.

\subsection{Cooperative\_treasure\_collection Scenario}
Cooperative\_treasure\_collection is a more complex competitive scenario constructed by \cite{iqbal2019actor} based on the framework of MPE. In this scenario, there are two types of agents, $X$ of which are treasure collectors, $Y$ of which are treasure banks. There are also $X$ treasures matching with corresponding color of the bank.
The role of collector aims to collect the treasures of any color and transport the treasures to the bank of the corresponding color. The treasures will be reborn after being collected and the banks just simply gather as much treasures as possible from collectors. 
When the treasures are collected by collectors, the collectors would share a global reward. 
At the same time, while the treasure is transported into the bank, all agents receive a global reward. However, collectors will be also penalized for collision with each other. To conclude, the collectors need to learn how to cooperatively collect treasures and deposit them into correct color bank as quick as possible without collision with other agents. The banks need to cooperate with collectors in placing treasures.
\begin{figure}
\centering
\subfigure[Tasks.]{\label{task3}
\includegraphics[width=0.47\textwidth]{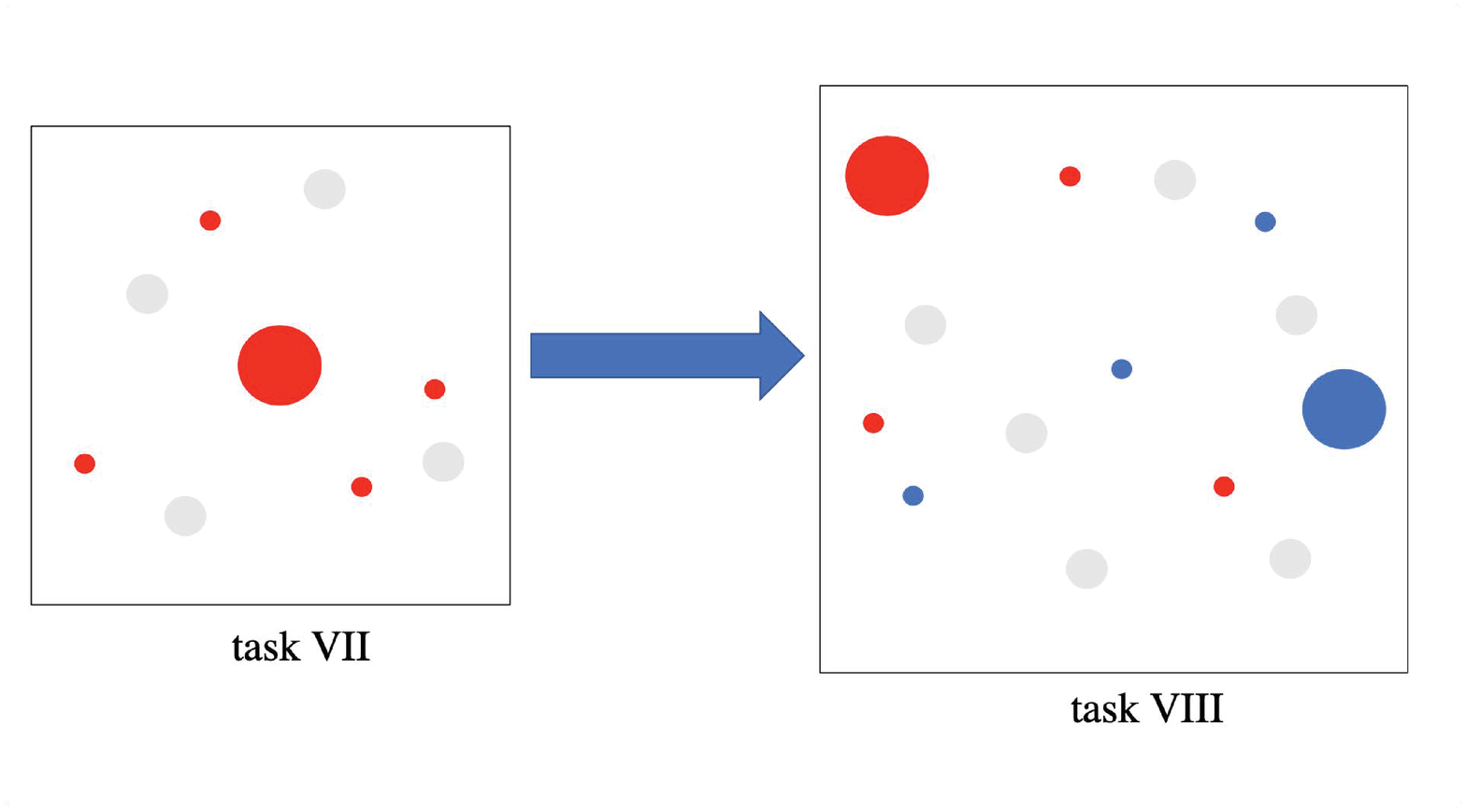}}
\hspace{0.01\linewidth}
\subfigure[Knowledge reusing in task VIII.]{\label{626}
\includegraphics[width=0.47\textwidth]{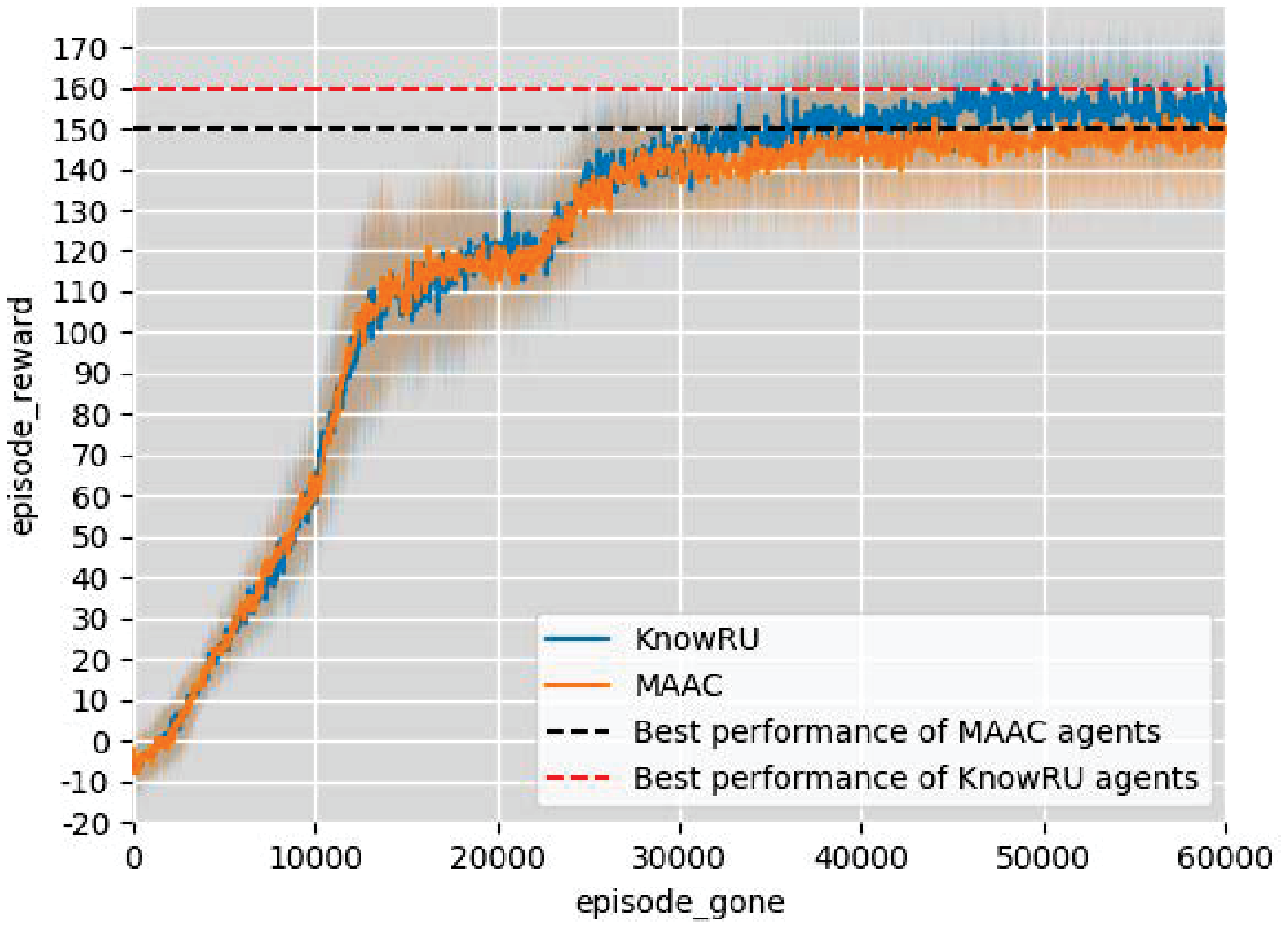}}
\caption{The tasks and results in Cooperative\_treasure\_collection.}
\end{figure}

In our experimental settings, we take the task which contains 4 collectors, one bank and 4 treasures as the teacher models’ training scenario, called task VII. And we design the scenario which is constituted of 6 collectors, 2 banks and 6 treasures as student models’ training scenarios, called task VIII, as shown in Figure~\ref{task3}. To test the universality, we use MAAC as the basic algorithm which can learn these tasks well with attention mechanism and implement KnowRU based on it. The actor networks in experiments are randomly parameterized by a four-layer fully connected MLP with 64 units per layer and critic networks are parameterized by attention mechanism. We set 60,000 episodes to ensure the convergence, every episode consist of 100 steps and the networks update 4 times every episode. The hyperparameter $\alpha$ and $i$ are set to 0.5 and 0.02. We take the average episode reward of all agents from environments to measure the effect of training. The higher the reward, the better. The results and 95\% confidence intervals (95\% CI) are illustrated in Figure~\ref{626}.

\textbf{Discussion.} As shown in Figure~\ref{626}, we complete the goals of time to threshold and asymptotic performance. MAAC firstly get and stabilize the reward value of 150 at about 37,000 episodes. However, KnowRU firstly achieve that at about 29,000 episodes and advance this time by about 21.6\% compared to MAAC. Compared to other two scenarios, we can draw some different conclusions and conjectures from this scenario. Even the two tasks in this scenario are not close, KnowRU also successfully help agents avoid falling into local optima and each a higher performance. The result proves our conjecture that KnowRU can help and guide the training of agents by providing more useful information.

\subsection{Component Analysis and Discussion}
\begin{figure}[h]
  \centering
  \includegraphics[width=\linewidth,height=6cm]{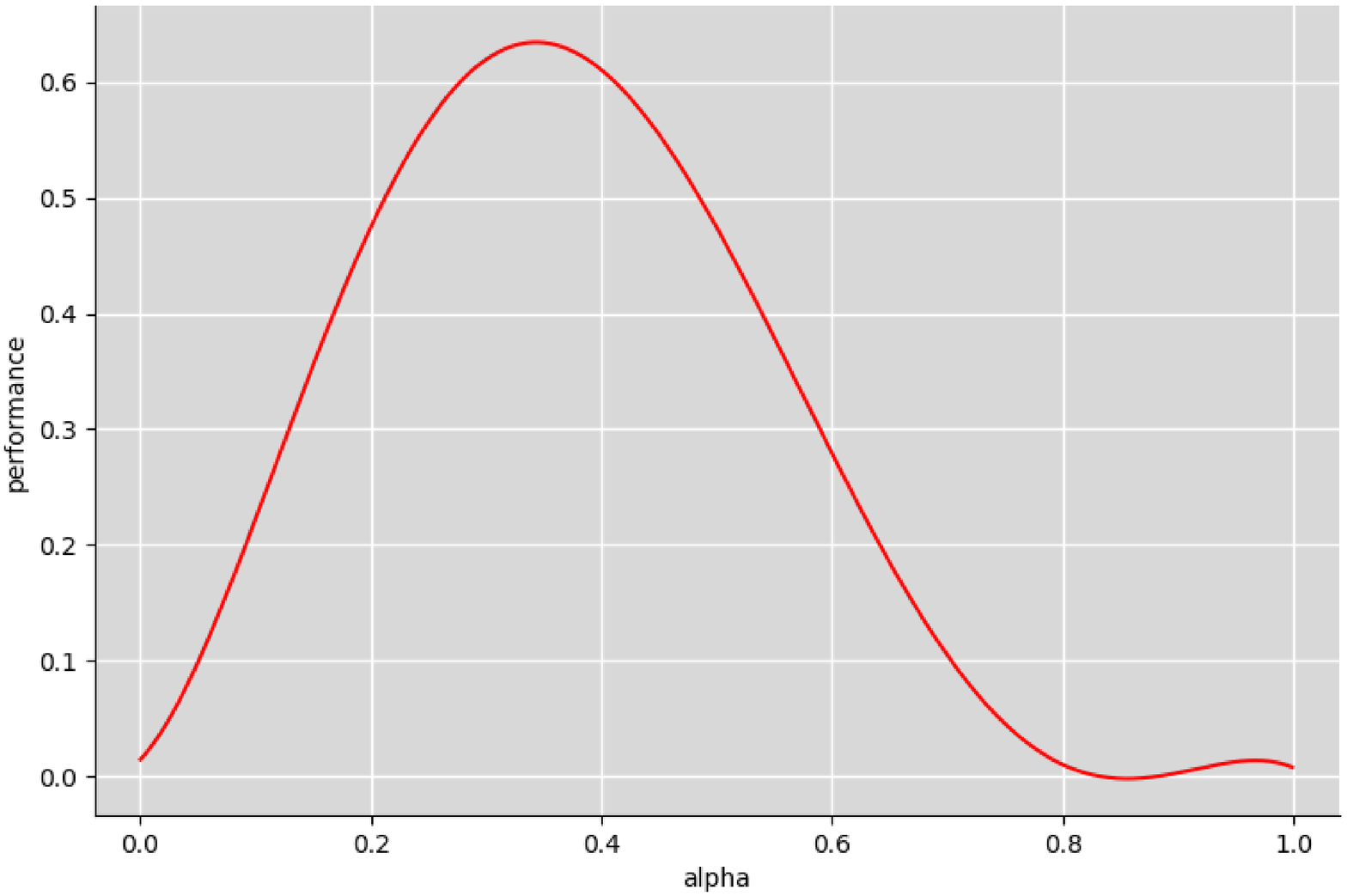}
  \caption{alpha.}
  \label{alpha}
\end{figure}

\textbf{Alpha.} We have declared that annealing of the alpha parameter to blend the guide phase and the specialization phase. We want to explore the impact of alpha on performance with Equation~(\ref{T}) and Equation~(\ref{per}).
\begin{equation}\label{T}
T=\left\{T_{\alpha 1}, T_{\alpha 2}, T_{\alpha 3} \cdots T_{\alpha n}\right\}
\end{equation}
\begin{equation}\label{per}
\text {Performance}_{\alpha}=\ln \left(\frac{\operatorname{Tmax}}{T \alpha}\right)
\end{equation}
where $T_{\alpha}$ represents the moment when the best performance is reached with $\alpha$, $T$ is the set of $T_{\alpha}$, Tmax is the maximum value of the set T and Performance$_{\alpha}$ denotes the performance of $\alpha$. We test the performance in task III of simple\_spread scenario and the result is shown in Figure~\ref{alpha}.
It should be pointed out that our method always obviously performs better than baseline when alpha is greater than 0.1. Here, we are just exploring the best settings for alpha. It's obvious that when the alpha value is around 0.3$\sim$0.5, the training performance is the best,  which is in line with our expectation that the training process is first guided and then specialized. We have also found similar patterns in other tasks. 

\textbf{Loss Function.} There are three viable loss functions for $\mathcal{L}_{reuse}$ we have tested, Mean-Square Error (MSE) loss, Cross-Entropy (CE) loss or Kullback-Leibler (KL) loss. The logits can be converted into probabilities with softmax function, then, probabilities are used in CE or KL. We find that using different loss functions did not have a significant impact on the experimental results. The experimental results are placed in additional materials.

\textbf{Discussion.} We have tried different combinations of components. The results did not show a huge difference and they all showed that KnowRU successfully help agents quickly adapt to the environment, which also proved the effectiveness and robustness of our method.

\section{Conclusion}
Transfer the knowledge from the historical experiences is of extensive practical interest for the MARL fields, yet notoriously unstable and difficult. In this paper, we explored a novel knowledge transfer approach for MARL and addresses its accompanying unique challenges, leveraging knowledge distillation paradigm. To empirically demonstrate the robustness and effectiveness of KnowRU, we perform extensive experiments on state-of-the-art MARL algorithms on collaborative and competitive scenarios. The results demonstrate the effectiveness and robustness of KnowRU under different experimental settings.

%
%

\bibliographystyle{splncs04} 
\bibliography{ref}
\end{document}